\documentclass{article}

\usepackage{arxiv}
\usepackage[numbers,sort&compress]{natbib}

\usepackage[utf8]{inputenc} 
\usepackage[T1]{fontenc}    
\usepackage{hyperref}       
\usepackage{url}            
\usepackage{booktabs}       
\usepackage{amsfonts}       
\usepackage{nicefrac}       
\usepackage{microtype}      
\usepackage{lipsum}
\usepackage{graphicx}
\usepackage{amsmath} 
\usepackage{algorithm}
\usepackage{caption}
\usepackage{pgfplots}
\usepackage{tikz}

\usetikzlibrary{positioning,calc,decorations.pathreplacing}

\usepackage{pgfplotstable}
\pgfplotsset{compat=1.18}
\usepackage[noend]{algpseudocode} 

\usepackage[utf8]{inputenc}

\usepackage{algpseudocode} 

\graphicspath{ {./images/} }

\title{Ranking of Bangla Word Graph using Graph-based Ranking Algorithms \\[0.5em]
\large Published in 2017 3rd International Conference on Electrical Information and Communication Technology (EICT), 7-9 December 2017, Khulna, Bangladesh}

\author{
S M Rafiuddin \\
Department of Computer Science and Engineering \\
Bangladesh University of Engineering and Technology \\
\texttt{torifat.cs@gmail.com} \\
}

\begin{document}
\maketitle
\begin{abstract}
Ranking words is an important way to summarize a text or to retrieve information. A word graph is a way to represent the words of a sentence or a text as the vertices of a graph and to show the relationship among the words. It is also useful to determine the relative importance of a word among the words in the word-graph. In this research, the ranking of Bangla words are calculated, representing Bangla words from a text in a word graph using various graph based ranking algorithms. There is a lack of a standard Bangla word database. In this research, the Indian Language POS-tag Corpora is used, which has a rich collection of Bangla words in the form of sentences with their parts of speech tags. For applying a word graph to various graph based ranking algorithms, several standard procedures are applied. The preprocessing steps are done in every word graph and then applied to graph based ranking algorithms to make a comparison among these algorithms. This paper illustrate the entire procedure of calculating the ranking of Bangla words, including the construction of the word graph from text. Experimental result analysis on real data reveals the accuracy of each ranking algorithm in terms of F1 measure.
\end{abstract}


\section{INTRODUCTION}

Natural Language Processing for a machine is an AI-complete problem. To represent Natural Language to a machine or a computer, a graph is an important data structure. Various algorithms that are used for Natural Language Processing has the input in a notion of a chart, which is a kind of graph. Word is the fundamental building block in Natural Language. Ranking word is a measurement of text classification, page ranking, text summarization, and various other applications. There are two types of ranking procedures. One is statistical methods andthe other is using graph based methods~\cite{FrakesBaeza1992}. In this research, several graph based ranking methods are applied to Bangla words, which are represented by a word graph data structure, and then the ranking accuracy is measured.

There are huge amount of unstructured data floating around the World Wide Web (WWW) that needs to be processed. To find the relevant terms and to extract the required information, we need to find the right terms associated with that required information. To do this, the relevant terms need to be ranked. However extracting the relevant terms manually is an extremely time-consuming process. So, to extract the relevant information from various documents, first we need to extract the relevant terms and form a word graph of these terms, then rank the relevant terms using graph based ranking algorithms.

Graph based ranking algorithms that are used to measure the accuracy of word graphs are HITS~\cite{Kleinberg1999}, positional power function~\cite{HeringsLaanTalman2001},and  PageRank algorithms~\cite{Page1999}. In the recent past, many researchers has used graph theory based algorithms to analyze Natural Language Processing. Although, graph theory and computational linguistics are two distinct fields of Computer Science, but some recent research work shows that graph theory can be very useful and sometimes connected to various aspects of Natural Language Processing.

Another importance of this research work is to apply these ranking algorithms to Bangla words that are represented via a word graph. This technique can be applied to a Bangla word based search engine. In this context of Bangla word ranking, there are scarcity of standard datasets. So, in this research Indian Language POS-tag corpora dataset is used for the ranking procedure.

So, the steps of the procedure of word ranking from different documents using the Bangla word graph are as follows—  
\begin{itemize}
  \item Read Documents
  \item Tokenization of Texts
  \item Stemming
  \item Stop Word deletion
  \item Build Word Graph
  \item Apply graph based ranking algorithms to word graphs
  \item Measure ranking accuracy
\end{itemize}

Here, Stemming means find the root word from the derived words, and stop word deletion means remove the irrelevant words that are not necessary for the consideration of ranking.

Word graphs are graph based syntactic representation of words and texts. The graphs are directed, acyclic where edges represents acoustic relations among the words. The word graph contains a unique start vertex and end vertex. These properties follow that the word graph is loosely connected. Edge labels denote words and weights denote acoustic numbers~\cite{OerderNey1993}.

\section{LITERATURE REVIEW}

Rada Mihalcea (2004) represents an unsupervised method for automatic text extraction for summarization process. In this research graph based ranking algorithms are applied to rank the words for summarization process~\cite{Mihalcea2004}. This paper investigates a range of standard algorithms and evaluates their applications to text summarization.

Florian Boudin (2013) presents various centrality measures for keyphrase extraction using graph based methods. Keyphrase of a text documents means words or phrases that precisely represent the associated document~\cite{Boudin2013}. Also this paper compares centrality measures of the experiments and results with related works of the recent past.

Litvak and Last (2008) introduce supervised and unsupervised approach of keyword extraction for single document. This method used graph based syntactic representation of text that enhances traditional vector space model. They also used HITS algorithm assuming that top ranked nodes of word graph represents most relevant terms of words~\cite{LitvakLast2008}. Then they represent their keyword identification accuracy using F1 score.

Khan et al. (2016) proposed an approach to extract domain relevant terms from an unstructured corpus. Their paper emphasized on minimizing the impact of term frequency to improve the precision of top k-terms. They also used unstructured data for categorization and information retrieval~\cite{KhanMaKim2016}. This paper also estimated term or word similarities based on term embedding.

Shams and Haratizadeh (2017) used collaborative ranking in graph based representation for item recommendation. They calculated users similarities based on ranking data. In this paper, a graph based approach called GRank is proposed and recommendation quality is compared with other collaborative ranking techniques~\cite{ShamsHaratizadeh2017}.

Sidiriopoulos and Manolopoulos (2006) used graph based ranking algorithms for publications and authors using citations and worked on the impact of scientific collections and authors~\cite{SidiropoulosManolopoulos2006}.

\section{METHODOLOGIES}

Graph based word ranking from a word graph is a procedure for deciding which vertex that represent the words with ranking that are relatively more relevant to the text. Let \(G\) be a graph. An edge \(e\) is called incident from \(u\) and incident to \(v\). For a vertex \(v\) of the graph \(G\), the in-degree is defined as number of edges incident to \(v\)~\cite{Amtrup1999}:
\begin{equation}
\mathrm{in\_deg}(v) = \lvert \{\,e \mid \text{incident to } v\}\rvert
\end{equation}

The out-degree of a vertex \(v\) is defined as the number of edges incident from \(v\)~\cite{Amtrup1999}:
\begin{equation}
\mathrm{out\_deg}(v) = \lvert \{\,e \mid \text{incident from } v\}\rvert
\end{equation}

Here, as access function to the start vertex of an edge defined as~\cite{Amtrup1999}:
\begin{equation}
\mathrm{src}(e) = u
\end{equation}

as access function to the end vertex of an edge defined as~\cite{Amtrup1999}:
\begin{equation}
\mathrm{tgt}(e) = v
\end{equation}

A word graph is a directed, acyclic graph \(G = (V,E)\) with edge labels and edge weights, if the following conditions hold~\cite{Amtrup1999}:
\begin{equation}
\forall e\in E:\ \text{src}(e),\,\text{tgt}(e)\in V
\end{equation}
\begin{equation}
\text{no directed cycles in }G
\end{equation}

Here, as access function to the label of an edge defined as~\cite{Amtrup1999}:
\begin{equation}
\ell(e) = \text{label of }e
\end{equation}

And, as access function to the weight of an edge defined as~\cite{Amtrup1999}:
\begin{equation}
w(e) = \text{weight of }e
\end{equation}

Applying topological ordering to Bangla word graph, as the word graph is a directed acyclic graph that has the mapping defined as~\cite{Amtrup1999}:
\begin{equation}
\pi : V \to \{1,\dots,|V|\}
\end{equation}

\begin{algorithm}
  \caption{Computing the topological order of word graph}
  \begin{algorithmic}[1]
    \State Initialize an empty stack \(s\).
    \ForAll{vertex \(v\)}
      \State mark \(v\) unvisited
      \State \(s.\mathrm{Push}(v)\)
    \EndFor
    \State \(\mathrm{order} \leftarrow \) nil
    \While{\(s\) not empty}
      \State \(v \leftarrow s.\mathrm{Pop}()\)
      \ForAll{edge \(e\) starting in \(v\)}
        \If{\(\mathrm{tgt}(e)\) unvisited}
          \State \(s.\mathrm{Push}(\mathrm{tgt}(e))\)
        \EndIf
      \EndFor
      \State append \(v\) to \(\mathrm{order}\)
    \EndWhile
    \State \Return \(\mathrm{order}\)
  \end{algorithmic}
\end{algorithm}

\begin{algorithm}
  \caption{Computation of the transcript independent density of a word graph}
  \begin{algorithmic}[1]
    \State \(\mathrm{dens} \leftarrow 0\)
    \State \(\mathrm{totalDens} \leftarrow 0\)
    \ForAll{vertex \(v\) in topological order}
      \State \(\mathrm{totalDens} \leftarrow \mathrm{totalDens} + \mathrm{dens}\)
      \State \(\mathrm{dens} \leftarrow \mathrm{dens} - \mathrm{in\_deg}(v)\)
    \EndFor
    \State \Return \(\mathrm{totalDens} / |V|\)
  \end{algorithmic}
\end{algorithm}

\begin{algorithm}
  \caption{Calculation of the number of paths in a word graph}
  \begin{algorithmic}[1]
    \ForAll{vertex \(v\) in topological order}
      \State \(\mathrm{paths}[v] \leftarrow 0\)
      \If{\(v\) is start vertex}
        \State \(\mathrm{paths}[v] \leftarrow 1\)
      \EndIf
      \ForAll{vertex \(u\) ending in \(v\)}
        \State \(\mathrm{paths}[v] \leftarrow \mathrm{paths}[v] + \mathrm{paths}[u]\)
      \EndFor
    \EndFor
    \State \Return \(\mathrm{paths}[\text{end vertex}]\)
  \end{algorithmic}
\end{algorithm}

\begin{algorithm}
  \caption{Reducing a graph to unique label sequences}
  \begin{algorithmic}[1]
    \ForAll{vertex \(v\) in topological order}
      \ForAll{pairs of identically labeled edges \(e_1,e_2\)}
        \State create new vertex \(v'\) and insert into topological order
        \ForAll{edge \((u \to v, e_1)\)}
          \State create edge \((u \to v')\)
        \EndFor
        \ForAll{edge \((v \to w, e_2)\)}
          \State create edge \((v' \to w)\)
        \EndFor
      \EndFor
    \EndFor
  \end{algorithmic}
\end{algorithm}

\begin{algorithm}
  \caption{Merging two vertices}
  \begin{algorithmic}[1]
    \ForAll{edge \(e\) incoming to \(v_2\)}
      \If{no parallel edge from \(\mathrm{src}(e)\) to \(v_1\)}
        \State create edge \((\mathrm{src}(e) \to v_1)\)
      \EndIf
    \EndFor
    \ForAll{edge \(e\) outgoing from \(v_2\)}
      \If{no parallel edge from \(v_1\) to \(\mathrm{tgt}(e)\)}
        \State create edge \((v_1 \to \mathrm{tgt}(e))\)
      \EndIf
    \EndFor
  \end{algorithmic}
\end{algorithm}

\begin{algorithm}
  \caption{Calculation of the number of steps of a fictitious parser}
  \begin{algorithmic}[1]
    \State \(\mathrm{totalderiv} \leftarrow 0\)
    \ForAll{vertex \(v\) in topological order}
      \State \(\mathrm{deriv}_v[1] \leftarrow 1\)
      \For{\(i = 2\) \textbf{to} \(|V|\)}
        \ForAll{edge \((u \to v)\)}
          \State \(\mathrm{deriv}_v[i] \leftarrow \mathrm{deriv}_v[i] + \mathrm{deriv}_u[i-1]\)
        \EndFor
        \State \(\mathrm{totalderiv} \leftarrow \mathrm{totalderiv} + \mathrm{deriv}_v[i]\)
      \EndFor
    \EndFor
    \State \Return \(\mathrm{totalderiv}\)
  \end{algorithmic}
\end{algorithm}

\begin{algorithm}
  \caption{Computing rank using minimum and maximum prefix scores}
  \begin{algorithmic}[1]
    \ForAll{vertex \(v\) in topological order}
      \State \(\mathrm{minScore}(v) \leftarrow +\infty\), \(\mathrm{maxScore}(v) \leftarrow 0\)
      \ForAll{edge \((u \to v)\)}
        \State \(\mathrm{minScore}(v) \leftarrow \min(\mathrm{minScore}(u) + w(u\to v),\ \mathrm{minScore}(v))\)
        \State \(\mathrm{maxScore}(v) \leftarrow \max(\mathrm{maxScore}(u) + w(u\to v),\ \mathrm{maxScore}(v))\)
      \EndFor
    \EndFor
    \State \Return \(\mathrm{minScore}, \mathrm{maxScore}\)
  \end{algorithmic}
\end{algorithm}

\begin{algorithm}
  \caption{Computing rank if the reference score is lower than or equal to the minimum score}
  \begin{algorithmic}[1]
    \If{\(\mathrm{refScore} \le \mathrm{minScore}(v)\)}
      \State \Return 0
    \ElsIf{\(\mathrm{refScore} \ge \mathrm{maxScore}(v)\)}
      \State \Return 1
    \EndIf
    \State \(\mathrm{score} \leftarrow 0\)
    \ForAll{edge \((u \to v)\)}
      \State \(\mathrm{score} \leftarrow \mathrm{score} + (\text{scoreContribution})\)
    \EndFor
    \State \Return \(\mathrm{score}\)
  \end{algorithmic}
\end{algorithm}

\begin{figure}[ht]
  \centering
  \begin{tikzpicture}[>=stealth]
    \node[draw,circle,inner sep=1.5pt] (n1) at (0,0) {};          
    \node[draw,circle,fill,inner sep=1.5pt] (n2) at (3,0) {};     
    \node[draw,circle,fill,inner sep=1.5pt] (n3) at (6,0) {};     
    \node[draw,circle,fill,inner sep=1.5pt] (n4) at (9,0) {};     
    \node[draw,circle,fill,inner sep=1.5pt] (n5) at (12,0) {};    
    \node[draw,circle,inner sep=1.5pt] (n6) at (15,0) {};        

    \draw[bend left=60] (n1) to node[midway, above] {0.5} (n2);
    \draw[bend left=30] (n1) to node[midway, above] {0.5} (n2);
    \draw                (n1) to node[midway, above] {0.0} (n2);
    \draw[bend right=30] (n1) to node[midway, below] {-0.5} (n2);
    \draw[bend right=60] (n1) to node[midway, below] {-0.5} (n2);

    \draw[bend left=60] (n2) to node[midway, above] {0.25} (n3);
    \draw[bend left=30] (n2) to node[midway, above] {0.25} (n3);
    \draw                (n2) to node[midway, above] {0.0} (n3);
    \draw[bend right=30] (n2) to node[midway, below] {-0.25} (n3);
    \draw[bend right=60] (n2) to node[midway, below] {-0.25} (n3);

    \draw[bend left=60] (n3) to node[midway, above] {0.125} (n4);
    \draw[bend left=30] (n3) to node[midway, above] {0.125} (n4);
    \draw                (n3) to node[midway, above] {0.0} (n4);
    \draw[bend right=30] (n3) to node[midway, below] {-0.125} (n4);
    \draw[bend right=60] (n3) to node[midway, below] {-0.125} (n4);

    \draw[bend left=60] (n4) to node[midway, above] {0.125} (n5);
    \draw[bend left=30] (n4) to node[midway, above] {0.125} (n5);
    \draw                (n4) to node[midway, above] {0.0} (n5);
    \draw[bend right=30] (n4) to node[midway, below] {-0.125} (n5);
    \draw[bend right=60] (n4) to node[midway, below] {-0.125} (n5);

    \draw[bend left=60] (n5) to node[midway, above] {0.125} (n6);
    \draw[bend left=30] (n5) to node[midway, above] {0.125} (n6);
    \draw                (n5) to node[midway, above] {0.0} (n6);
    \draw[bend right=30] (n5) to node[midway, below] {-0.125} (n6);
    \draw[bend right=60] (n5) to node[midway, below] {-0.125} (n6);
  \end{tikzpicture}
  \caption{Word-graph edges with successive link weights.}
  \label{fig:word-graph-successive}
\end{figure}
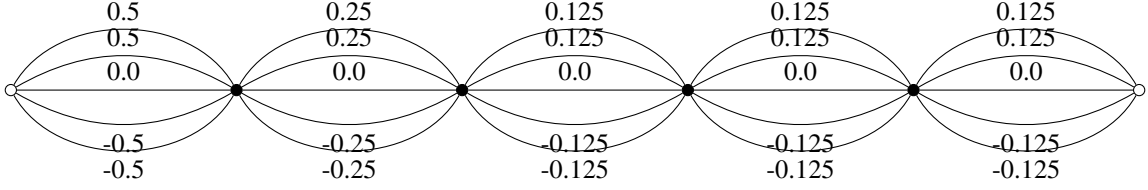

\subsection*{Algorithm 9: Hyperlinked Induced Topic Search (HITS)~\cite{Kleinberg1999}}

This is an iterative algorithm developed by Kleinberg in 1999~\cite{Kleinberg1999}. It ranks web pages using authority measurements. Authority is measured using the number of incoming links. This algorithm recursively assigns a weight to each vertex determining how relevant the word graph vertex is. For each vertex in a word graph we calculate two scores: authority and hub. Calculating the ranking of a vertex in a word graph using HITS is an unsupervised approach.

Authority score is defined as
\begin{equation}
\mathrm{auth}(v_j) \;=\; \sum_{\,i:\,(v_i\to v_j)\in E} \mathrm{hub}(v_i)
\label{eq:auth}
\end{equation}
Hub score is defined as
\begin{equation}
\mathrm{hub}(v_i) \;=\; \sum_{\,j:\,(v_i\to v_j)\in E} \mathrm{auth}(v_j)
\label{eq:hub}
\end{equation}

To derive a single rank \(R(v)\) from these two scores, we consider:
\begin{enumerate}
  \item Rank associated with the authority score:
    \begin{equation}
      R_1(v_j) \;=\; \mathrm{auth}(v_j)
    \end{equation}
  \item Rank associated with the hub score:
    \begin{equation}
      R_2(v_j) \;=\; \mathrm{hub}(v_j)
    \end{equation}
  \item Rank associated with the average of both scores:
    \begin{equation}
      R_3(v_j) \;=\; \tfrac{1}{2}\bigl(\mathrm{auth}(v_j) + \mathrm{hub}(v_j)\bigr)
    \end{equation}
  \item Rank associated with the maximum of both scores:
    \begin{equation}
      R_4(v_j) \;=\; \max\{\mathrm{auth}(v_j), \mathrm{hub}(v_j)\}
    \end{equation}
\end{enumerate}

\begin{algorithm}
  \caption{HITS Iteration}
  \begin{algorithmic}[1]
    \State \textbf{Input:} \(G=(V,E)\)
    \ForAll{\(v\in V\)} 
      \State \(\mathrm{auth}(v)\leftarrow1\), \(\mathrm{hub}(v)\leftarrow1\)
    \EndFor
    \Repeat
      \ForAll{\(v\in V\)}
        \State \(\mathrm{auth}(v)\leftarrow \sum_{u:(u\to v)\in E}\mathrm{hub}(u)\)
      \EndFor
      \ForAll{\(v\in V\)}
        \State \(\mathrm{hub}(v)\leftarrow \sum_{u:(v\to u)\in E}\mathrm{auth}(u)\)
      \EndFor
      \State Normalize \(\{\mathrm{auth}(v)\},\;\{\mathrm{hub}(v)\}\)
    \Until{convergence}
    \State \Return \(\{\mathrm{auth}(v)\}\)
  \end{algorithmic}
\end{algorithm}

\subsection*{Algorithm 10: Positional Power Function (PPF)~\cite{HeringsLaanTalman2001}}

This ranking algorithm determines the ranking of a vertex as a function of the number and score of its successors. Positional Power Function is defined as:
\begin{equation}
\mathrm{PPF}(v_j) \;=\; \sum_{\,p=(v_0\to\cdots\to v_k=v_j)} 
\sum_{i=1}^{k}\frac{w(v_{i-1}\to v_i)}{i}
\label{eq:ppf}
\end{equation}

\begin{algorithm}
  \caption{PPF Computation}
  \begin{algorithmic}[1]
    \State \textbf{Input:} \(G=(V,E)\)
    \ForAll{\(v\in V\)}
      \State \(\mathrm{PPF}(v)\leftarrow 0\)
      \ForAll{paths \(p=(v_0\to\cdots\to v_k=v)\)}
        \For{\(i=1\) \textbf{to} \(k\)}
          \State \(\mathrm{PPF}(v)\leftarrow \mathrm{PPF}(v) + \tfrac{w(v_{i-1}\to v_i)}{i}\)
        \EndFor
      \EndFor
    \EndFor
    \State \Return \(\{\mathrm{PPF}(v)\}\)
  \end{algorithmic}
\end{algorithm}

\subsection*{Algorithm 11: PageRank~\cite{Page1999}}

This ranking algorithm integrates both incoming and outgoing links. It is defined as:
\begin{equation}
\mathrm{PR}(v_j) \;=\; \frac{1-d}{|V|} 
\;+\; d \sum_{\,i:(v_i\to v_j)\in E} \frac{\mathrm{PR}(v_i)}{\mathrm{out\_deg}(v_i)}
\label{eq:pagerank}
\end{equation}
where \(d\in(0,1)\) is the damping factor. Values are updated until convergence.

\begin{algorithm}
  \caption{PageRank Iteration}
  \begin{algorithmic}[1]
    \State \textbf{Input:} \(G=(V,E)\), damping \(d\)
    \ForAll{\(v\in V\)} 
      \State \(\mathrm{PR}(v)\leftarrow \tfrac1{|V|}\)
    \EndFor
    \Repeat
      \ForAll{\(v\in V\)}
        \State \(\mathrm{PR}(v)\leftarrow \tfrac{1-d}{|V|} 
               + d \sum_{u:(u\to v)\in E}\tfrac{\mathrm{PR}(u)}{\mathrm{out\_deg}(u)}\)
      \EndFor
      \State Normalize \(\{\mathrm{PR}(v)\}\)
    \Until{convergence}
    \State \Return \(\{\mathrm{PR}(v)\}\)
  \end{algorithmic}
\end{algorithm}

\section{EXPERIMENTS AND RESULTS}

\subsection{A. Dataset Description}

The experiment of this research project is Indian Language POS-tag corpus~\cite{NLTKCorpus}. The corpora has several language e.g. Bangla, Hindi, Marathi and Telegu. This research is working with \texttt{bangla.pos} file inside of this corpora. This database is completely publicly available and accessible via Natural Language Processing Toolkit. The database is compiled with XML file and has 895 Bangla sentence.

\subsection{B. Experimental Setup}

\paragraph{Hardware Setup:}
\begin{itemize}
  \item Processor: AMD X4 Phenom II 965 BE 3.4GHz
  \item RAM: 8GB DDR3 1333MHz
  \item HDD: 1TB
\end{itemize}

\paragraph{Software Setup:}
\begin{itemize}
  \item Operating System: Linux Mint XFCE 18.2
  \item Programming Language: Python Programming Language
  \item Version: Python 3.4.1
  \item Used packages: Natural Language Toolkit (NLTK)
\end{itemize}

\subsection{C. Results}

The metric of accuracy has four distinct features~\cite{MishuRafiuddin2016}. They are defined as follows—
\begin{itemize}
  \item True positives are relevant terms that are correctly ranked.
  \item True negatives are irrelevant terms that are correctly ranked.
  \item False positives are irrelevant terms that are incorrectly ranked.
  \item False negatives are relevant terms that are incorrectly ranked.
\end{itemize}

By these four features, two parameters of F measure is calculated. First parameter i.e. Precision, is calculated as how many terms that are ranked properly, i.e.:
\[
\text{Precision} = \frac{\text{True Positives}}{\text{True Positives} + \text{False Positives}}
\]
Second parameter i.e. Recall, is calculated as how many relevant terms that are identified, i.e.:
\[
\text{Recall} = \frac{\text{True Positives}}{\text{True Positives} + \text{False Negatives}}
\]
From Precision and Recall, the metric of accuracy, i.e. F1 is calculated as:
\[
\mathrm{F1} = 2 \times \frac{\text{Precision} \times \text{Recall}}{\text{Precision} + \text{Recall}}
\]

Using F1 score, the accuracy of ranking from word graph derived from Indian Language POS-tag Corpora, by graph based ranking algorithms are presented below:

\begin{table}[ht]
  \centering
  \caption{Result Analysis of Ranking Algorithms from Bangla Word Graph}
  \label{tab:results}
  \begin{tabular}{lrrrr}
    \toprule
    \textbf{Ranking Algorithm} & \textbf{\# Words} & \textbf{Precision} & \textbf{Recall} & \textbf{F1 Score} \\
    \midrule
    Rank using minimum and maximum prefix & 1234 & 0.36 & 0.40 & 0.37 \\
    Rank considering reference score      & 1259 & 0.32 & 0.38 & 0.33 \\
    HITS                                   & 1435 & 0.31 & 0.37 & 0.32 \\
    Positional Power Function             & 1278 & 0.30 & 0.35 & 0.31 \\
    PageRank                               & 1546 & 0.29 & 0.36 & 0.31 \\
    \bottomrule
  \end{tabular}
\end{table}

The following figure depicts the ranking algorithm accuracy in a column graph:

\begin{figure}[ht]
  \centering
  \begin{tikzpicture}
    \begin{axis}[
      ybar,
      bar width=8pt,
      enlarge x limits=0.25,
      ylabel={Score},
      symbolic x coords={
        Min/Max Prefix,
        Reference Score,
        HITS,
        PPF,
        PageRank
      },
      xtick=data,
      x tick label style={rotate=35,anchor=east},
      ymin=0, ymax=0.45,
      ymajorgrids,
      grid style=dashed,
      legend style={
        at={(0.5,-0.25)},
        anchor=north,
        legend columns=3
      }
    ]
      \addplot+[fill=blue] coordinates {
        (Min/Max Prefix,0.36)
        (Reference Score,0.32)
        (HITS,0.31)
        (PPF,0.30)
        (PageRank,0.29)
      };
      \addplot+[fill=red] coordinates {
        (Min/Max Prefix,0.40)
        (Reference Score,0.38)
        (HITS,0.37)
        (PPF,0.35)
        (PageRank,0.36)
      };
      \addplot+[fill=green] coordinates {
        (Min/Max Prefix,0.37)
        (Reference Score,0.33)
        (HITS,0.32)
        (PPF,0.31)
        (PageRank,0.31)
      };

      \legend{Precision, Recall, F1 Score}
    \end{axis}
  \end{tikzpicture}
  \caption{Precision, Recall and F1 Scores of Graph-Based Ranking Algorithms}
  \label{fig:results-multi-bar}
\end{figure}
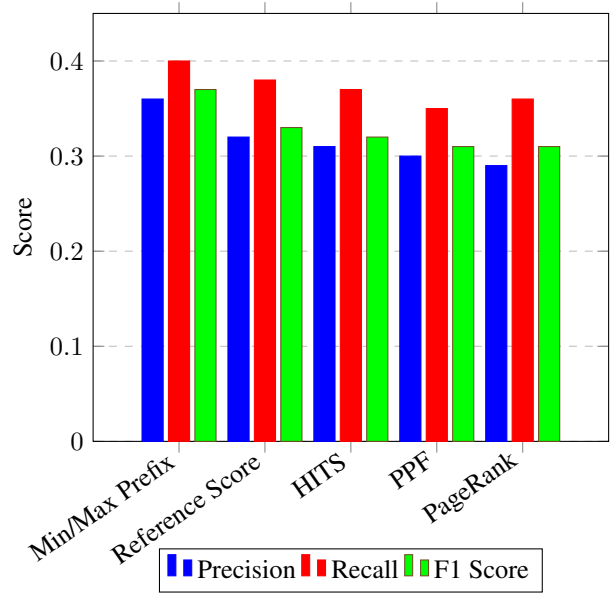

\section{CONCLUSION}

This research has the goal of accuracy of the ranking algorithms of Bangla words using graph-based ranking algorithms. Bangla is the 7th most spoken language in the world. The text containing Bangla words are increasing at a very high rate in newspapers, blogs, and social media. So, searching the documents containing Bangla words are getting important in the recent research field. For searching, summarizing, and information retrieval of Bangla text, ranking Bangla word using graph based ranking algorithms are important. So, Bangla word ranking will grow more attention in Natural Language Processing field of Bangla Language. There are a scarcity of Bangla word databases. This research has used Indian Language POS-tag Corpora for Bangla words. To represent the words in a data structure, using graph word is very useful. So, to process Bangla words, several word-graph methods and techniques are used to process the work in the text that has been discussed in the methodologies portion. The future work of this research is to optimize the ranking algorithm up to a certain level and to improve these algorithm for the Bangla word ranking process.

\section*{ACKNOWLEDGMENT}

The author(s) would like to thank Professor Dr. Mohammad Kaykobad, Department of Computer Science and Engineering (CSE), Bangladesh University of Engineering and Technology (BUET), for his kind support and enthusiasm. Also, this research work is very much grateful to Dr. Sarada Herke, Postdoctoral Research Fellow, School of Mathematics and Physics, University of Queensland, Australia, for her tutorials on Graph Theory on YouTube.

\end{document}